\documentclass[conference]{IEEEtran}

\usepackage{amsmath,amssymb,amsfonts}
\usepackage{algorithmic,latexsym}
\usepackage{graphicx}
\usepackage{textcomp}
\usepackage{url,stackrel}
\usepackage{bm,float}
\usepackage{xcolor,tabu}

\usepackage{epsfig}
\usepackage{epstopdf}
\usepackage{stfloats}
\usepackage{color,empheq}
\usepackage{booktabs}
\usepackage{textgreek}

\makeatletter
\renewcommand*\env@matrix[1][\arraystretch]{%
  \edef\arraystretch{#1}%
  \hskip -\arraycolsep
  \let\@ifnextchar\new@ifnextchar
  \array{*\c@MaxMatrixCols c}}
\makeatother

\begin{document}

\title{Analyzing Uncertainty in the Spatial Representation of the Kinematic Bicycle Model}



\author{\IEEEauthorblockN{Shafayat Abrar}
\IEEEauthorblockA{Dhanani School of\\ Science and Engineering\\
Habib University, Karachi 75290\\
Email: shafayat.abrar@sse.habib.edu.pk}\\ 
\IEEEauthorblockN{Shahir Ul Islam Anzal}
\IEEEauthorblockA{Dhanani School of\\ Science and Engineering\\
Habib University, Karachi 75290\\
Email: sa06081@alumni.habib.edu.pk}\\
\and
\IEEEauthorblockN{M. Zaeem Baig}
\IEEEauthorblockA{Dhanani School of\\ Science and Engineering\\
Habib University, Karachi 75290\\
Email: mb05962@alumni.habib.edu.pk}\\
\IEEEauthorblockN{Abdul Basit Memon}
\IEEEauthorblockA{Dhanani School of\\ Science and Engineering\\
Habib University, Karachi 75290\\
Email: basit.memon@sse.habib.edu.pk}}

\maketitle

\begin{abstract}
Locating a vehicle and determining its orientation in an uncertain environment is a critical challenge in autonomous vehicle navigation and path planning. To address these challenges, a vehicle estimates its pose while depending on sensor data that offer noisy measurements. These uncertainties in pose quantities are expressed mathematically as a covariance matrix. The real-time computation of the covariance matrix is critical because of the non-linearity involved in the kinematic model. The challenge is thus to evaluate the evolution of the covariance matrix of a vehicle's discretized stochastic kinematics. 

The purpose of this study is to obtain a near-accurate evolution of the covariance matrix of the rear-wheel bicycle kinematic model under uncertainties in wheel displacement and steering angle. We used Taylor's series to linearize the nonlinear trigonometric functions and provided closed-form expectations of random variables with the required accuracy. Our analytical findings are in good agreement with those obtained from Monte-Carlo simulations. 
Our contribution is probably the first detailed closed-form presentation of covariance matrix constituents of the vehicle under evaluation, which were previously reported either incorrectly or incompletely. These findings aid in identifying the potential and constraints of the discretized kinematic model as well as its stochastic analysis. The techniques presented here are useful to the simultaneous localization and odometry self-calibration of certain mobile robots and autonomous vehicles.
\end{abstract}


\section{Introduction}
Autonomous mobile robots (AMRs) are widely used across industries such as manufacturing, logistics, healthcare, agriculture, security, and search and rescue. Their success is largely due to advanced navigation systems that enable efficient and safe movement while avoiding obstacles. This paper contributes to the field of localization, which involves estimating a robot’s position and orientation, known as its pose, within an environment. Accurate localization is crucial for task execution, obstacle avoidance, and goal attainment.  

A robot’s pose can be updated at each time step using odometric methods, which rely on motion data and prior pose estimates. However, even well-engineered robots exhibit odometry errors, categorized as systematic and non-systematic errors. Systematic errors arise from design flaws, mechanical imperfections, or sensor inaccuracies, while non-systematic errors depend on the environment. Systematic errors are relatively constant over short intervals and can be corrected through calibration techniques, such as the ``UMBmark'' method introduced by Borenstein et al. \cite{Borenstein1996}. Non-systematic errors, in contrast, are random and cannot be eliminated, making them particularly significant in environments with limited sensor feedback. These errors are often modeled as random variables with parametric distributions, introducing uncertainty in pose estimation, which is the focus of this paper.  

The first approach to quantifying pose uncertainty was proposed by Brooks \cite{brooks1985visual}, representing it as min/max position bounds that were updated as the robot moved. A similar approach was introduced in \cite{chatila1985position}. Later, covariance matrices became the standard method for representing pose uncertainty, as demonstrated in \cite{RandallCheeseman1986, SmithandCheeseman}. These matrices quantify pose uncertainty at a given time and are updated at each time step.  

Updating the covariance matrix generally requires knowledge of the random error model and the odometry motion model, which describes how a robot’s pose evolves based on internal motion measurements. However, the nonlinear nature of odometry models makes covariance propagation complex. To address this, a Jacobian-based linearization approach was proposed in \cite{RandallCheeseman1986, SmithandCheeseman}, assuming the availability of one-step linear and angular displacements and following a Gaussian error model. The validity of this method was analytically confirmed in \cite{MingWangJ1989}, where it was shown to be effective for small orientation changes and displacements between updates.  

More recently, \cite{FilhoAccess2019} analyzed the impact of parametric uncertainties in a mobile robot’s kinematic model on velocity and pose estimation. The study introduced models that quantify these uncertainties, considering factors like wheel radius and inter-wheel distance. Simulation-based experiments validated the models, highlighting their significance for localization, control, and robotic system design.

This paper aims to enhance the accuracy of covariance update methods, building on prior efforts in the literature. The first significant work was by Wang \cite{MingWangC1988, MingWangJ1990}, who introduced three alternative approximations to Jacobian-based linearization for updating the one-step covariance matrix. These were derived for a differential-drive kinematic model, assuming known left and right wheel displacements and a Gaussian error model. While Wang’s method improved pose estimates for larger angular displacements, it assumed motion along a circular path of constant radius between updates.  

Kleeman and Chong \cite{Kleeman1995, ChongKleeman1996} divided a differential-drive robot’s path into $N$ smaller segments, obtaining closed-form covariance update expressions as $N$ approached infinity, assuming small angular displacements per segment. Their work focused on three fundamental paths: constant-radius circular arcs, straight motion, and rotation about the wheel axis, arguing that these can approximate any arbitrary path.  

Kelly \cite{kelly2003general, KellyAlonzo2004} later proposed a comprehensive covariance propagation solution applicable to any error model, odometry model, and robot path, though it still relied on Jacobian-based linearization. Martinelli \cite{martinelli2002odometry} advanced the field by eliminating this approximation but only for synchronous drive systems, assuming errors depend solely on incremental path length, not angular error.

This paper builds on the work of Tur et al. \cite{Borja2005, Borja2007, Borja2009} and proposes improved closed-form covariance update expressions for an Ackerman-drive robot, modeled as a rear-wheel bicycle. Their earlier study \cite{Borja2005} omitted the cross-covariance between the previous pose estimate and current pose error, which was later addressed in \cite{Borja2007}. This paper retains the cross-covariance terms while improving accuracy by refining approximations for the expected value of the tangent of the steering angle.  

The paper is structured as follows: Section~\ref{sec:model} describes the robot’s kinematic model, Section~\ref{sec:covariance} details the computation of covariance update terms, and Section~\ref{sec:simulation} compares the proposed expressions against Monte Carlo simulations and the results of Tur et al. \cite{Borja2007}. 

\section{Kinematic Model of Rear-Wheel Bicycle}
\label{sec:model}
The pose of a mobile robot, i.e. its position and orientation, in 2D space is represented by the triplet $(x,y,\theta)$. For a rear-wheel bicycle, the external kinematic model
may be expressed as a set of following  differential equations \cite{Klancar2017}: 
\[
\begin{aligned}
&\dot{x}(t)=v(t) \cos(\theta(t)) \\
 &\dot{y}(t)=v(t) \sin(\theta(t)) \\
 &\dot{\theta}(t)=\omega(t)=\textstyle\frac{1}{L}v(t)\tan(\phi(t))
\end{aligned}
\]
where $(x,y,\theta)$ are global pose quantities, $v$ is linear velocity, $\phi$ is steering angle, and $L$ is the length of bicycle. Assuming $v$ and $\phi$ are control/input variables, the vehicle pose for given control/input variables can be determined through direct (or forward) kinematics:
$$
\begin{aligned}
& x(t)=x\left(t_0\right)+\textstyle\int_{t_0}^t v(\tau) \cos (\theta(\tau)) d \tau \\
& y(t)=y\left(t_0\right)+\textstyle\int_{t_0}^t v(\tau) \sin (\theta(\tau)) d \tau \\
& \theta(t)=\theta\left(t_0\right)+\textstyle\frac{1}{L}\displaystyle\textstyle\int_{t_0}^t v(\tau)\tan(\phi(\tau)) d \tau
\end{aligned}
$$
Defining a discrete time, $k$, such that $x_k$ represents the value of $x(t)$ at $t=k T$, where $T$ is the sampling time, we obtain
$$
\begin{aligned}
x_{k+1}&=x_{k}+\textstyle\int_{k T}^{(k+1) T}v(t) \cos (\theta(t)) d t\\
y_{k+1}&=y_{k}+\textstyle\int_{k T}^{(k+1) T}v(t) \sin (\theta(t)) d t\\
\theta_{k+1}&=\theta_k+\textstyle\frac{1}{L}\displaystyle\textstyle\int_{k T}^{(k+1) T} v(t)\tan(\phi(t)) d t
\end{aligned}
$$
Note that the velocity and steering angle, the control inputs, are assumed fixed in the time interval \(t\in[kT,(k+1)T)\); therefore, \(v_k:=v(t)|_{t=kT}\) and \(\phi_k:=\phi(t)|_{t=kT}\) are fixed for \(t\in[kT,(k+1)T)\). With this consideration, we  obtain
\begin{equation}
\label{EqKinematicTheta}
\theta_{k+1}=\theta_k+\frac{\Delta d_k}{L} \tan(\phi_k)=:\theta_k+\Delta\theta_k
\end{equation}
where \(\Delta d_k:=Tv_k\) is the linear distance covered by the wheel in time $T$, \(\Delta\theta_k=\Delta d_k\tan(\phi_k)/L=:T\omega_k\) is the angular displacement of steering wheel from its mean position, 
and $\omega_k$ denotes the turn rate (angular frequency), and consequently the ratio $v_k/\omega_k$ describes the instantaneous radius of curvature. Further, 
the so-called \textit{exact integration method} \cite[Eq. (2.7)]{Klancar2017} helps us obtain:
\begin{subequations}\label{EqKinematicXY}
\begin{alignat}{2}
& x_{k+1}=x_k+\frac{v_k}{\omega_k}\left(\sin \left(\theta_k+\Delta\theta_k \right)-\sin (\theta_k)\right) \\
& y_{k+1}=y_k-\frac{v_k}{\omega_k}\left(\cos \left(\theta_k+\Delta\theta_k \right)-\cos (\theta_k)\right)
\end{alignat}    
\end{subequations}
In addition, as described in \cite{MingWangJ1990} and \cite{ColinDas2010}, using geometric and trigonometric arguments, respectively, an alternate representation of the kinematics in (\ref{EqKinematicXY}) may be obtained as follows:
\begin{subequations}\label{EqKinematicXY2}
\begin{alignat}{2}
& x_{k+1}=x_k+\Delta d_k\operatorname{sinc} \left(\tfrac{1}{2}\Delta\theta_k\right)\cos \left(\theta_k+\tfrac{1}{2}\Delta\theta_k\right)=:& ~x_k+\Delta {x}_k \\
& y_{k+1}=y_k+\Delta d_k\operatorname{sinc} \left(\tfrac{1}{2}\Delta\theta_k\right)\sin \left(\theta_k+\tfrac{1}{2}\Delta\theta_k\right)=:& ~y_k+\Delta {y}_k
\end{alignat}    
\end{subequations}
where $\operatorname{sinc}(\cdot)$ is an unnormalized sinc function which is defined as \(\operatorname{sinc}(x)=\sin(x)/x\).\footnote{The sinc factor in Eq. (\ref{EqKinematicXY2}) has been termed as \textit{adjustment factor} in the literature; it was first coined by Ming Wang \cite{MingWangJ1986,MingWangJ1987} who studied discretization of curvilinear motion of autonomous vehicles. This and similar factors come into action when variables of a continuous-time system of curvilinear motion are transformed into those of discrete-time. More sinc function-based factors come into play when motion with acceleration, jerk, snap, etc., is discretized. For a motion in a straight line, however, all such factors become unity (because the turn rate is zero). Interested readers may refer to \cite{Langholm2021} for sinc function-based factors in kinematic models in polar coordinate systems.} The equations in (\ref{EqKinematicXY2}) and (\ref{EqKinematicTheta}) will be adopted as the kinematic model for the rear-wheel bicycle in this paper. Note that the kinematics in (\ref{EqKinematicXY2}) do not assume any translation in the lateral direction. \\

The  wheel distance, $\Delta d_k$, and the steering angle, $\phi_k$ are considered as control/inputs in this model, but these are obtained from imperfect odometric measurements. The imperfection in odometric measurements arises from sensor noise, the roughness of the terrain, and wheel slippage. These odometric errors are modeled as random variables. Let $ {\widehat{\Delta d}}_k$ and $\widehat{\phi}_k$ denote, respectively, the  wheel distance and steering angle obtained from odometric measurements at time index $k$. So, the measurements $ {\widehat{\Delta d}}_k$ and $\widehat{\phi}_k$ contain true values $\Delta{d}_k$ and $\phi_k$, respectively, added with zero-mean random errors, $\widetilde{\Delta {d}}_k$ and $\widetilde{\phi}_k$. Thus, at time index $k$, we have: 
\begin{subequations}
\begin{align}
     {\widehat{\Delta d}}_k&=\Delta d_k+\widetilde{\Delta {d}}_k \\
\widehat{\phi}_k&=\phi_k+\widetilde{\phi}_k \end{align}
\end{subequations}
where we assume that \(\widetilde{\Delta {d}}_k\) and \(\widetilde{\phi}_k\) are independent of each other, and normally distributed as follows: 
\[\widetilde{\Delta {d}}_k \sim \mathcal{N}\left(0, \sigma_{\widetilde{\Delta d}_k}^2\right)~~ \text{and}~~\widetilde{\phi}_k \sim \mathcal{N}\left(0, \sigma_{\widetilde{\phi}_k}^2\right)~~\forall k.\]
For the sake of analysis or convenience, the variances $\sigma_{\widetilde{\Delta d}_k}^2$ and $\sigma_{\widetilde{\phi}_k}^2$ are sometimes considered fixed; this is; however, not true in practice.

Using the kinematic model, (\ref{EqKinematicXY2}) and (\ref{EqKinematicTheta}), the pose of the vehicle can be determined from the measured quantities by the following expressions:
\begin{subequations}\label{EqPoseWithErrors}
\begin{align}
\nonumber\widehat{x}_{k+1}&=\widehat{x}_{k}+\widehat{\Delta x}_k \\ &=\widehat{x}_{k}+ {\widehat{\Delta d}}_k \operatorname{sinc} \left(\tfrac{1}{2}\widehat{\Delta\theta}_k\right) \cos \left(\widehat{\theta}_{k}+\textstyle\tfrac{1}{2}{\widehat{\Delta \theta}_k}\right) \\
\nonumber\widehat{y}_{k+1}&=\widehat{y}_{k}+\widehat{\Delta y}_k\\ &=\widehat{y}_{k}+ {\widehat{\Delta d}}_k \operatorname{sinc} \left(\tfrac{1}{2}\widehat{\Delta\theta}_k\right) \sin \left(\widehat{\theta}_{k}+\textstyle\tfrac{1}{2}{\widehat{\Delta \theta}_k}\right) \\
\widehat{\theta}_{k+1}&=\widehat{\theta}_{k}+\widehat{\Delta \theta}_k =\widehat{\theta}_{k}+\textstyle\frac{1}{L}{ {\widehat{\Delta d}}_k \tan \widehat{\phi}_k}
\end{align}
\end{subequations}

In literature, the statistical analysis of recursions (\ref{EqPoseWithErrors}) is done exploiting one of four different approaches. The most popular and the  simplest approach assumes that the adjustment factor, $\operatorname{sinc} (\tfrac{1}{2}\widehat{\Delta\theta}_k)$, is \textit{unity} which is largely true if the change in orientation angle is small because \(\operatorname{sinc}(\theta)/\theta\rightarrow 1\) as \(\theta\rightarrow 0\). The second approach replaces $\operatorname{sinc} (\tfrac{1}{2}\widehat{\Delta\theta}_k)$ with its deterministic value $\operatorname{sinc} (\tfrac{1}{2}\Delta\theta_k)$ and consider it a fixed factor for the given time index $k$. The third approach exploits the following manipulations using Taylor's series approximation up to the second-order:
\begin{align*}
    &\operatorname{sinc} \left(\tfrac{1}{2}\widehat{\Delta\theta}_k\right) \cos \left(\widehat{\theta}_{k}+\textstyle\tfrac{1}{2}{\widehat{\Delta \theta}_k}\right)\\
    & ~~=\frac{1}{\widehat{\Delta\theta}_k}\left[\sin \left(\widehat{\theta}_{k}+{\widehat{\Delta \theta}_k}\right)
    - \sin\widehat{\theta}_{k}\right]\\ &~~\approx \frac{1}{\widehat{\Delta\theta}_k}\left[\sin\widehat{\theta}_{k}+{\widehat{\Delta \theta}_k}\cos\widehat{\theta}_{k}-\tfrac{1}{2}\left({\widehat{\Delta \theta}_k}\right)^2\sin\widehat{\theta}_{k}
    - \sin\widehat{\theta}_{k}\right]\\
    &~~= \cos \widehat{\theta}_{k}-\tfrac{1}{2}\widehat{\Delta \theta}_k\sin \widehat{\theta}_{k}
\end{align*}
The fourth approach tends to deal with the factor in statistical analysis in a near exact manner. In this work, we choose to stick with the first approach where the $\operatorname{sinc}$ factor is replaced with unity. So, the reduced/simplified kinematic modeled considered in this work is given by:
\begin{subequations}\label{EqPoseWithErrorsSimplified}
\begin{align}
\widehat{x}_{k+1} &=\widehat{x}_{k}+ {\widehat{\Delta d}}_k \cos \left(\widehat{\theta}_{k}+\textstyle\tfrac{1}{2}{\widehat{\Delta \theta}_k}\right) \\
\widehat{y}_{k+1} &=\widehat{y}_{k}+ {\widehat{\Delta d}}_k  \sin \left(\widehat{\theta}_{k}+\textstyle\tfrac{1}{2}{\widehat{\Delta \theta}_k}\right) \\
\widehat{\theta}_{k+1}&=\widehat{\theta}_{k}+\widehat{\Delta \theta}_k =\widehat{\theta}_{k}+\textstyle\frac{1}{L}{ {\widehat{\Delta d}}_k \tan \widehat{\phi}_k}
\end{align}
\end{subequations}

This is interesting to note that three incremented states, $\widehat{x}_{k+1}$, $\widehat{y}_{k+1}$, and $\widehat{\theta}_{k+1}$, depend only on one prior estimate, i.e., $\widehat{\theta}_{k}$, where $\widehat{\theta}_{k}$  consists of the true orientation ${\theta}_{k}$ and the estimation error $\widetilde{\theta}_{k}$, then, at a given time index $k$, we have
\begin{equation}
\widehat{\theta}=\theta+\widetilde{\theta},
\end{equation}
where \(\widetilde{\theta} \sim \mathcal{N}\big(0, \text{var}(\widetilde{\theta}_{k})\big)\); so, it is assumed that at the given time index $k$, $\widehat{\theta}$ is normally distributed with mean zero, and variance $\sigma_{\widetilde{\theta}}^2$. This is important to realize that the variance $\text{var}(\widetilde{\theta}_{k})$ is not fixed and tends to grow with time as the vehicle moves in its uncertain environment; this is explained below:
\begin{equation}\label{eqvariance}
\begin{aligned}
\text{var}(\widetilde{\theta}_{k})&=\text{var}(\widehat{\theta}_{k})=\text{var}(\widehat{\theta}_{k-1}+\widehat{\Delta\theta}_{k})\\
& ~~~~~\vdots\\
& =\text{var}(\widehat{\theta}_{1}+\widehat{\Delta\theta}_{2}+\cdots+\widehat{\Delta\theta}_{k-1}+\widehat{\Delta\theta}_{k})\\
& =\text{var}(\widehat{\theta}_{0}+\widehat{\Delta\theta}_{1}+\widehat{\Delta\theta}_{2}+\cdots+\widehat{\Delta\theta}_{k-1}+\widehat{\Delta\theta}_{k})
\end{aligned}
\end{equation}
Note that \(\widehat{\Delta \theta}_k = \frac{1}{L} \widehat{\Delta d}_k \tan \widehat{\phi}_k\), where the measured quantities \(\widehat{\Delta d}_k\) and \(\widehat{\phi}_k\) contain random components, \(\varepsilon_{\Delta{d}}\) and \(\varepsilon_{{\phi}}\), which are independent and uncorrelated. Based on this, we can reasonably assume that \(\widehat{\Delta\theta}_i\) and \(\widehat{\Delta\theta}_j\) are also independent and uncorrelated for \(i \neq j\)\footnote{As discussed in \cite[Sec. 3.2]{RandallCheeseman1986}, \cite{SmithandCheeseman}, covariance analysis relies on three key assumptions: (a) the first-order approximation, where the Jacobian of the transformation evaluated at the mean values of the variables is sufficiently \textit{accurate}, (b) the errors in the transformations being combined are \textit{independent}, and (c) these errors have \textit{zero mean}.}. Consequently, (\ref{eqvariance}) simplifies to
\begin{equation}
\text{var}(\widetilde{\theta}_{k})=\sum_{i=1}^k\text{var}(\widehat{\Delta\theta}_i)
\end{equation}
Since, $\sum_{i=1}^{k-1}\text{var}(\widehat{\Delta\theta}_i)$ represents $\text{var}(\widetilde{\theta}_{k-1})$, we may write
\begin{equation}\label{EqMonotonicity}
\text{var}(\widetilde{\theta}_{k})=\text{var}(\widetilde{\theta}_{k-1})+\text{var}(\widehat{\Delta\theta}_k).
\end{equation}
Equation (\ref{EqMonotonicity}) demonstrates \textit{monotonicity} that the uncertainty increases during the course of vehicle movement; interested readers must refer to \cite{RodriguezArevalo2018, KnuthPrabir2013} for a detailed discussion on the monotonicity in the propagation of spatial uncertainties. 

\section{Covariance Analysis}
\label{sec:covariance}
The uncertainty of the entire pose estimate is represented by its covariance matrix, which is computed in this section. Let the instantaneous increment in the pose vector be given by \(\Delta \widehat{\mathbf{p}}_k=\big[
\widehat{\Delta x}_k, 
\widehat{\Delta y}_k, 
\widehat{\Delta \theta}_k\big]^\prime\). 
The covariance of $\Delta \widehat{\mathbf{p}}_k$ is given as follows:
\begin{align}
\operatorname{cov}\left(\Delta \widehat{\mathbf{p}}_k,\Delta \widehat{\mathbf{p}}_k\right)=\mathbf{E}\left[\Delta \widehat{\mathbf{p}}_k \Delta \widehat{\mathbf{p}}_k^\prime\right]-\mathbf{E}\left[\Delta \widehat{\mathbf{p}}_k\right] \mathbf{E}\left[\Delta \widehat{\mathbf{p}}_k^\prime\right]
\end{align}
where $\mathbf{E}[\cdot]$ denotes the expected value of the quantity of interest. 
The two terms in the covariance are expressed as:
\begin{align}
\mathbf{E}[\Delta \widehat{\mathbf{p}}_k \Delta \widehat{\mathbf{p}}_k^\prime]=\begin{pmatrix}[1.5]
\mathbf{E}[\widehat{\Delta x}_k^2] & \mathbf{E}[\widehat{\Delta x}_k \widehat{\Delta y}_k] & \mathbf{E}[\widehat{\Delta x}_k \widehat{\Delta \theta}_k] \\
 \mathbf{E}[\widehat{\Delta y}_k \widehat{\Delta x}_k] & \mathbf{E}[\widehat{\Delta y}_k^2] & \mathbf{E}[\widehat{\Delta y}_k \widehat{\Delta \theta}_k] \\
\mathbf{E}[\widehat{\Delta \theta}_k \widehat{\Delta x}_k] & \mathbf{E}[\widehat{\Delta \theta}_k \widehat{\Delta y}_k] & \mathbf{E}[\widehat{\Delta \theta}_k^2]
\end{pmatrix}
\end{align}
and
\begin{align}
&\nonumber\mathbf{E}[\Delta \widehat{\mathbf{p}}_k] \mathbf{E}[\Delta \widehat{\mathbf{p}}_k^\prime]=\\ 
&\begin{pmatrix}[1.5]
\mathbf{E}[\widehat{\Delta x}_k]^2 & \mathbf{E}[\widehat{\Delta x}_k] \mathbf{E}[\widehat{\Delta y}_k] & \mathbf{E}[\widehat{\Delta x}_k] \mathbf{E}[\widehat{\Delta \theta}_k] \\
\mathbf{E}[\widehat{\Delta y}_k] \mathbf{E}[\widehat{\Delta x}_k] & \mathbf{E}[\widehat{\Delta y}_k]^2 & \mathbf{E}[\widehat{\Delta y}_k] \mathbf{E}[\widehat{\Delta \theta}_k] \\
\mathbf{E}[\widehat{\Delta \theta}_k] \mathbf{E}[\widehat{\Delta x}_k] & \mathbf{E}[\widehat{\Delta \theta}_k] \mathbf{E}[\widehat{\Delta y}_k] & \mathbf{E}[\widehat{\Delta \theta}_k]^2
\end{pmatrix}
\end{align}
where $\mathbf{E}[z]^2$ and $\mathbf{E}[z^2]$ denote the square of the expected value and the expected value of the square of variable $z$, respectively.

\subsection{Evaluation of the variance of orientation angle}

The uncertainty in the measured value of orientation depends directly on the uncertainties in the wheel displacement and steering angle.   
\begin{eqnarray}\label{eqvaianceoftheta}
\nonumber
& & \text{var}(\widehat{\Delta\theta}_k)=\text{cov}[\widehat{\Delta\theta}_k, \widehat{\Delta\theta}_k]=\mathbf{E}[\widehat{\Delta\theta}_k^2]-\mathbf{E}[\widehat{\Delta\theta}_k]^2\\ 
& & \nonumber=\tfrac{1}{L^2}\mathbf{E}[( {\widehat{\Delta d}}_k)^2 \tan^2(\widehat{\phi}_k)]-\tfrac{1}{L^2}\mathbf{E}[ {\widehat{\Delta d}}_k\tan(\widehat{\phi}_k)]^2
\\ & & \nonumber 
=\tfrac{1}{L^2}\mathbf{E}[( {\widehat{\Delta d}}_k)^2]\,\mathbf{E}[ \tan^2(\widehat{\phi}_k)]-\tfrac{1}{L^2}\mathbf{E}[ {\widehat{\Delta d}}_k]^2\,\mathbf{E}[\tan(\widehat{\phi}_k)]^2.\\
\end{eqnarray}
where the last equality is based on the independence of $\epsilon_{\Delta d}$ and $\widetilde{\phi}$. The evaluation of $\mathbf{E}\big[\big( {\widehat{\Delta d}}_k\big)^2\big]$ and $\mathbf{E}\big[ {\widehat{\Delta d}}_k\big]^2$ is straightforward; $\mathbf{E}\big[\big( {\widehat{\Delta d}}_k\big)^2\big]=\Delta {d}_k^2+\sigma_{\Delta d}^2$
and $\mathbf{E}\big[ {\widehat{\Delta d}}_k\big]^2=\Delta {d}_k^2$.
The evaluation of quantities involving $\widehat{\phi}_k$, however, needs attention.
A demonstrable approach is to exploit the Taylor series expansions of $\tan \big(\widehat{\phi}_k\big)$ and $\tan^2 \big(\widehat{\phi}_k\big)$ about $\tan(\widetilde{\phi}\big)=0$. Therefore
\begin{align}\label{eqTaylorseriesoftheta}
& \nonumber \tan \widehat{\phi}_k=\tan (\phi_k+\widetilde{\phi})\\ \nonumber & =\tan\phi_k+\sec^2(\phi_k)\textstyle\sum_{n=1}^{\infty} \tan^n(\widetilde{\phi}\,) \tan^{n-1}(\phi_k)\\
& \nonumber=\tan\phi_k+\sec^2(\phi_k) \tan(\widetilde{\phi}\,) +\tan\phi_k\sec^2(\phi_k) \tan^2(\widetilde{\phi}\,) \\
& \nonumber +\tan^2(\phi_k)\sec^2(\phi_k) \tan^3(\widetilde{\phi}\,) +\tan^3(\phi_k)\sec^2(\phi_k) \tan^4(\widetilde{\phi}\,)\\ & +O(\tan^5(\widetilde{\phi}\,) )
\end{align}
Now taking the expectations of both sides, we obtain:
\begin{align}\label{EqExpectationofDeltaTheta} \nonumber \mathbf{E}[\tan (\widehat{\phi}_k)]
& =\tan\phi_k+\tan\phi_k\sec^2(\phi_k)\mathbf{E}[\tan^2(\widetilde{\phi}\,) ] \\ & +\tan^3(\phi_k)\sec^2(\phi_k) \mathbf{E}[\tan^4(\widetilde{\phi}\,) ]
\end{align}
where we have assumed that expectations of orders higher than four are negligible.
Similarly, we obtain
\begin{align}
\label{eqTaylorseriesofthetaSquare}
\nonumber
\tan^2 (\phi_k+\widetilde{\phi})
\nonumber  & = \tan^2(\phi_k)\\ & \nonumber  \nonumber  \nonumber  +\sec^2(\phi_k)\textstyle\sum_{n=1}^{\infty} \tan^n(\widetilde{\phi}\,) \tan^{n-2}(\phi_k)\\ &\times(n-1+(1+n)\tan^2(\phi_k))
\nonumber\\ & =\tan^2(\phi_k)+2\sec^2(\phi_k)\tan\phi_k \tan(\widetilde{\phi}\,)\nonumber \\ & +\sec^2(\phi_k)(3 \tan^2(\phi_k)+1) \tan^2(\widetilde{\phi}\,) \nonumber \\ & +\,2\sec^2(\phi_k)\tan\phi_k(2 \tan^2(\phi_k)+1) \tan^3(\widetilde{\phi}\,) \nonumber\\ & +\tan^2(\phi_k)\sec^2(\phi_k)(5 \tan^2(\phi_k)+3) \tan^4(\widetilde{\phi}\,) \nonumber\\ & 
+O(\tan^5(\widetilde{\phi}\,) )\end{align}
Again, taking expectations of both sides, we obtain:
\begin{align}\label{EqExpectationofDeltaThetaSquare}
\nonumber
& \mathbf{E}[\tan^2 (\phi_k+\widetilde{\phi})]  \approx \tan^2(\phi_k)\\  \nonumber
 &
 \quad +\sec^2(\phi_k)(3 \tan^2(\phi_k)+1) \mathbf{E}[\tan^2(\widetilde{\phi}\,) ]  \\ &\quad +\tan^2(\phi_k)\sec^2(\phi_k)(5 \tan^2(\phi_k)+3) \mathbf{E}[\tan^4(\widetilde{\phi}\,) ]
\end{align}
where we have assumed that expectations of orders higher than four are negligible.

Next, we evaluate $\mathbf{E}\big[\tan^2\big(\widetilde{\phi}\,\big) \big]$ and $\mathbf{E}\big[\tan^4\big(\widetilde{\phi}\,\big) \big]$. Using Taylor's series expansion, we may show
\begin{subequations}\begin{alignat}{2}
& \tan^2\big(\widetilde{\phi}\,\big) =\widetilde{\phi}^2 + \tfrac{2}{3}\widetilde{\phi}^4 + \tfrac{17}{45} \widetilde{\phi}^6 + \tfrac{62}{315} \widetilde{\phi}^8 + O(\widetilde{\phi}^9),\\ 
&\tan^4\big(\widetilde{\phi}\,\big) =\widetilde{\phi}^4 + \tfrac{4}{3}\widetilde{\phi}^6 + \tfrac{6}{5} \widetilde{\phi}^8 + \tfrac{848}{945} \widetilde{\phi}^{10} + O(\widetilde{\phi}^{11}).\end{alignat}
\end{subequations}
Since, \(\widetilde{\phi} \sim \mathcal{N}(0, \sigma_\phi)\), we have  
\(\mathbf{E}\big[\widetilde{\phi}^n\big]=
\sigma_\phi^n (n-1)!!\) for even $n$; \(\mathbf{E}\big[\widetilde{\phi}^n\big]\) is zero for odd $n$;
therefore, 
\begin{subequations}\label{EqExpectationofDeltaPhi24}\begin{alignat}{2}
& \mathbf{E}\big[\!\tan^2\big(\widetilde{\phi}\,\big) \big]=\sigma_\phi^2 + 2\sigma_\phi^4 + \tfrac{17}{3} \sigma_\phi^6 + \tfrac{62}{3}\sigma_\phi^8 + O(\sigma_\phi^9),\\
& \mathbf{E}\big[\!\tan^4\big(\widetilde{\phi}\,\big) \big]=3\sigma_\phi^4 + 20\sigma_\phi^6 + 126 \sigma_\phi^8 + 848\sigma_\phi^{10} + O(\sigma_\phi^{11}).
\end{alignat}    
\end{subequations}
Using (\ref{EqExpectationofDeltaTheta}), (\ref{EqExpectationofDeltaThetaSquare}) and (\ref{EqExpectationofDeltaPhi24}), we obtain the variance of $\widehat{\Delta\theta}_k$.


\subsection{Evaluation of the variances of location coordinates}
\allowdisplaybreaks

For the given system, $\widehat{\Delta x}_k$ is given by \(\widehat{\Delta x}_k=\widehat{\Delta  d}_k \cos \left(\theta_{k}+\widetilde{\theta}_{k}+\tfrac{1}{2}\widehat{\Delta \theta}_k\right)\),
where \(\widetilde{\theta}_{k}\) is the estimation error. The squared $\widehat{\Delta x}_k$ is expanded as follows:
\begin{align}
\widehat{\Delta x}_k^2&= {\widehat{\Delta d}}_k^2 \cos^2(\theta_{k}+\widetilde{\theta}_{k}+\tfrac{1}{2}\widehat{\Delta \theta}_k)
\nonumber\\
& = {\widehat{\Delta d}}_k^2g[\cos^2(\tfrac{1}{2}\widehat{\Delta\theta}_{k}) \cos^2(\theta_{k}) \cos^2(\widetilde{\theta}_{k}) \nonumber\\& + \cos^2(\tfrac{1}{2}\widehat{\Delta\theta}_{k}) \sin^2(\theta_{k}) \sin^2(\widetilde{\theta}_{k}) \nonumber\\ & +\sin^2(\tfrac{1}{2}\widehat{\Delta\theta}_{k}) \cos^2(\theta_{k}) \sin^2(\widetilde{\theta}_{k})\nonumber\\ & +\sin^2(\tfrac{1}{2}\widehat{\Delta\theta}_{k}) \cos^2(\widetilde{\theta}_{k}) \sin^2(\theta_{k})\nonumber\\ & -
\tfrac{1}{2} \cos^2(\tfrac{1}{2}\widehat{\Delta\theta}_{k})\sin(2\theta_{k}) \sin(2\widetilde{\theta}_{k}) 
\nonumber\\ & + \tfrac{1}{2}
\sin^2(\tfrac{1}{2}\widehat{\Delta\theta}_{k})\sin(2\theta_{k}) \sin(2\widetilde{\theta}_{k})\nonumber \\ & 
- \tfrac{1}{2} \sin(\widehat{\Delta\theta}_{k}) \cos^2(\widetilde{\theta}_{k}) \sin(2\theta_{k}) \nonumber\\ & - \tfrac{1}{2} \sin(\widehat{\Delta\theta}_{k}) \cos^2(\theta_{k}) \sin(2\widetilde{\theta}_{k}) \nonumber\\ & + \tfrac{1}{2}  \sin(\widehat{\Delta\theta}_{k})\sin(2\theta_{k}) \sin^2(\widetilde{\theta}_{k})\nonumber\\ & + \tfrac{1}{2} \sin(\widehat{\Delta\theta}_{k})\sin^2(\theta_{k}) \sin(2\widetilde{\theta}_{k})]
\end{align}
Since $\Delta \theta_k$ is small, we may comfortable assume that \(\cos \left(\tfrac{1}{2}{\widehat{\Delta \theta}_k}\right) \approx 1\), and \(\sin \left(\tfrac{1}{2}\widehat{\Delta \theta}_k\right) \approx \tfrac{1}{2}\widehat{\Delta \theta}_k\). We obtained the following expression
\begin{align}
\widehat{\Delta x}_k^2&\approx  {\widehat{\Delta d}}_k^2[ \cos^2(\theta_{k}) \cos^2(\widetilde{\theta}_{k}) +  \sin^2(\theta_{k}) \sin^2(\widetilde{\theta}_{k}) \nonumber \\ & \!\!\!\! +(\tfrac{1}{2}\widehat{\Delta\theta}_{k})^2\cos^2(\theta_{k}) \sin^2(\widetilde{\theta}_{k}) +(\tfrac{1}{2}\widehat{\Delta\theta}_{k})^2 \cos^2(\widetilde{\theta}_{k}) \sin^2(\theta_{k}) \nonumber \\ & -
\tfrac{1}{2} \sin(2\theta_{k}) \sin(2\widetilde{\theta}_{k}) 
+ \tfrac{1}{2}
(\tfrac{1}{2}\widehat{\Delta\theta}_{k})^2\sin(2\theta_{k}) \sin(2\widetilde{\theta}_{k}) \nonumber \\ & 
- \tfrac{1}{2} (\widehat{\Delta\theta}_{k}) \cos^2(\widetilde{\theta}_{k}) \sin(2\theta_{k}) - \tfrac{1}{2} (\widehat{\Delta\theta}_{k}) \cos^2(\theta_{k}) \sin(2\widetilde{\theta}_{k}) \nonumber \\ & + \tfrac{1}{2}  (\widehat{\Delta\theta}_{k})\sin(2\theta_{k}) \sin^2(\widetilde{\theta}_{k}) + \tfrac{1}{2} (\widehat{\Delta\theta}_{k})\sin^2(\theta_{k}) \sin(2\widetilde{\theta}_{k})]\nonumber \\
= &~ {\widehat{\Delta d}}_k^2\{ \cos^2(\theta_{k}) \cos^2(\widetilde{\theta}_{k}) +  \sin^2(\theta_{k}) \sin^2(\widetilde{\theta}_{k}) \nonumber\\ & -
\tfrac{1}{2} \sin(2\theta_{k}) \sin(2\widetilde{\theta}_{k})\}  +\tfrac{ {\widehat{\Delta d}}_k^4\tan^2(\widehat{\phi}_k)}{4L^2}\{\cos^2(\theta_{k}) \sin^2(\widetilde{\theta}_{k})\nonumber\\ &  +\cos^2(\widetilde{\theta}_{k}) \sin^2(\theta_{k}) +\tfrac{1}{2}
\sin(2\theta_{k}) \sin(2\widetilde{\theta}_{k})\} \nonumber\\ &  + \tfrac{ {\widehat{\Delta d}}_k^3\tan(\widehat{\phi}_k)}{2L}\{\sin(2\theta_{k}) (\sin^2(\widetilde{\theta}_{k}) \nonumber -\cos^2(\widetilde{\theta}_{k}))\nonumber\\ &  + \sin(2\widetilde{\theta}_{k})(\sin^2(\theta_{k})
 - \cos^2(\theta_{k}))\}
\end{align}
We compute the expectation of sine and cosine arguments (of $\widetilde{\theta}_{k}$) by decomposing them into polynomial series using Taylor's method. Alternatively, we may also use moment-generating functions for such computations, refer to \cite{Chapman1970}. Since, \(\widetilde{\theta}_k\sim\mathcal{N}\left(0,\text{var}\big(\widetilde{\theta}_{k}\big)\right)\), we obtain  
\begin{subequations}
\begin{alignat}{4}
\mathbf{E}\big[\cos\big(\widetilde{\theta}_{k}\big)\big] &  =\mathbf{E}\left[\sum_{n=0}^{\infty} \frac{(-1)^n}{(2 n) !} \widetilde{\theta}_{k}^{\,2 n}\right]=\sum_{n=0}^{\infty} \frac{(-1)^n}{(2 n) !} \mathbf{E}\left[\widetilde{\theta}_{k}^{\,2 n}\right]\nonumber \\ 
& =\sum_{n=0}^{\infty} \frac{(-1)^n}{(2 n) !} (2 n-1)!!\,\Big(\text{var}\big(\widetilde{\theta}_{k}\big)\Big)^n\nonumber \\
& =\sum_{n=0}^{\infty} \frac{\left(-\text{var}\big(\widetilde{\theta}_{k}\big) / 2\right)^n}{n !}=\exp \left(-\text{var}\big(\widetilde{\theta}_{k}\big)/{2}\right)\\
    \mathbf{E}\big[\sin \big(\widetilde{\theta}_{k}\big)\big]&=\sum_{n=0}^{\infty} \frac{(-1)^n}{(2 n+1) !} \mathbf{E}\left[\widetilde{\theta}_{k}^{\,2 n+1}\right]\nonumber\\ & =\sum_{n=0}^{\infty} \frac{(-1)^n}{(2 n+1) !}(0)=0.
\end{alignat}\label{EqCosineSine}
\end{subequations}
Based on the above considerations and exploiting the independence between the uncertainty and estimation error, we obtain the expectation of $\widehat{\Delta x}_k^2$ as follows:

\begin{align}
\textbf{E}[\widehat{\Delta x}_k^2] \nonumber &  = ( c_{1} \cos^2(\theta_{k})+c_{2} \sin^2(\theta_{k}))\textbf{E}[\Delta{\widehat{d}}_k^2]\nonumber\\ & +\frac{(c_2-c_1)\sin(2\theta_{k})}{2L}\mathbf{E}[\tan (\widehat{\phi}_k)]\textbf{E}[\Delta{\widehat{d}}_k^3] \nonumber\\ 
&+\frac{c_{1} \sin^2(\theta_{k})+c_{2} \cos^2(\theta_{k})}{4 L^2}\mathbf{E}[\tan^2 (\widehat{\phi}_k)]\textbf{E}[\Delta{\widehat{d}}_k^4].
\end{align}

where 
\[\begin{aligned}
& c_1=\textbf{E}\big[\cos^2\big(\widetilde{\theta}_{k}\big)\big]=\tfrac{1}{2}+\tfrac{1}{2}\exp\big(-2\,\text{var}\big(\widetilde{\theta}_{k}\big)\big)\\
&c_2:=\textbf{E}\big[\sin^2\big(\widetilde{\theta}_{k}\big)\big]=\tfrac{1}{2}-\tfrac{1}{2}\exp\big(-2\,\text{var}\big(\widetilde{\theta}_{k}\big)\big)
\end{aligned}
\]
Similarly, we obtain
\begin{align}
\textbf{E}[\widehat{\Delta y}_k^2] & = ( c_{2} \cos^2(\theta_{k})+c_{1} \sin^2(\theta_{k}))\textbf{E}[\Delta{\widehat{d}}_k^2]\nonumber \\ &+\frac{(c_1-c_2)\sin(2\theta_{k})}{2L}\mathbf{E}[\tan (\widehat{\phi}_k)]\textbf{E}[\Delta{\widehat{d}}_k^3]\nonumber
\\ &
+\frac{c_2 \sin^2(\theta_{k})+c_1 \cos^2(\theta_{k})}{4 L^2}\mathbf{E}[\tan^2 (\widehat{\phi}_k)]\textbf{E}[\Delta{\widehat{d}}_k^4].
\end{align}

\subsection{Evaluation of the (cross) covariances}

The cross-covariances are obtained as follows:
\begin{align}
\widehat{\Delta x}_k \widehat{\Delta y}_k& = {\widehat{\Delta d}}_k^2 \cos (\widehat{\theta}_{k}+\tfrac{1}{2} \widehat{\Delta \theta}_k) \sin (\widehat{\theta}_{k}+\tfrac{1}{2} \Delta \widehat{\theta}_{k})\nonumber\\
& =\tfrac{1}{2}  {\widehat{\Delta d}}_k^2 \sin (2 \widehat{\theta}_{k}+\widehat{\Delta \theta}_k) \nonumber\\ & =\tfrac{1}{2}  {\widehat{\Delta d}}_k^2 \sin (2 \theta_{k}+2 \widetilde{\theta}_{k}+\widehat{\Delta \theta}_k)\nonumber\\
\mathbf{E}[\widehat{\Delta x}_k \widehat{\Delta y}_k]&=\tfrac{c_3}{2}\sin (2\theta_{k})  \mathbf{E}[ \widehat{\Delta d}_k^2] \nonumber \\ & +\tfrac{c_3}{2 L}\cos (2\theta_{k})   
\mathbf{E}[\tan \widehat{\phi}_k]
\mathbf{E}[\widehat{\Delta d}_k^3]
\end{align}
where \(c_3:=\mathbf{E}\big[\cos \big(2 \widetilde{\theta}_{k}\big)\big]=\exp\big(-2\,\text{var}\big(\widetilde{\theta}_{k}\big)\big)\).
\begin{align*}
&\widehat{\Delta x}_k \widehat{\Delta \theta}_k\nonumber = \textstyle\frac{1}{L}\widehat{\Delta d}_k^{\,2} \tan \widehat{\phi}_k
 \cos (\theta_{k}+ \widetilde{\theta}_{k}+\tfrac{1}{2} \widehat{\Delta \theta}_k)\nonumber\\ \nonumber
&= \frac{\widehat{\Delta d}_k^{\,2} \cos\theta_{k}\cos\widetilde{\theta}_{k} \tan\widehat{\phi}_k}{L}-  \frac{\widehat{\Delta d}_k^{\,3} \cos\widetilde{\theta}_{k} \sin\theta_{k} \tan^2(\widehat{\phi}_k)}{2 L^2} \nonumber \\ \nonumber &  -\frac{\widehat{\Delta d}_k^{\,3} \cos\theta_{k}\sin\widetilde{\theta}_{k} {\tan(\widehat{\phi}_k)}^2}{2 L^2}-\frac{\widehat{\Delta d}_k^{\,2} \sin\theta_{k}\sin\widetilde{\theta}_{k}\tan\widehat{\phi}_k}{L}\end{align*}
\begin{align}
\mathbf{E}[\widehat{\Delta x}_k \widehat{\Delta \theta}_k]&=\frac{c_4}{L}\cos(\theta_{k})\mathbf{E}[\tan (\widehat{\phi}_k)]\mathbf{E}[ \widehat{\Delta d}_k^{\,2}] \nonumber \\ &  -\frac{c_4}{2 L^2}\sin(\theta_{k})   
\mathbf{E}[\tan^2(\widehat{\phi}_k)]\mathbf{E}[\widehat{\Delta d}_k^{\,3}]
\end{align}
where \(c_4:=\mathbf{E}\big[\cos \big(\widetilde{\theta}_{k}\big)\big]\) as  in (\ref{EqCosineSine}a). Similarly, we obtain
\begin{align}
\mathbf{E}[\widehat{\Delta y}_k \widehat{\Delta \theta}_k] \nonumber & =\frac{c_4}{L}\sin\theta_{k}\mathbf{E}[\tan \widehat{\phi}_k]\mathbf{E}[\widehat{\Delta d}_k^{\,2}] \nonumber \\ & +\frac{c_4}{2 L^2}\cos \theta_{k}   
\mathbf{E}[\tan^2(\widehat{\phi}_k)]
\mathbf{E}[\widehat{\Delta d}_k^{\,3}].
\end{align}
Finally the first-order moments of $\widehat{\Delta x}_k$ and $\widehat{\Delta y}_k$ are obtained
\begin{align}
\widehat{\Delta x}_k&= {\widehat{\Delta d}}_k \cos(\theta_{k}+\widetilde{\theta}_{k}+\tfrac{1}{2}\widehat{\Delta \theta}_k) \nonumber\\
& = {\widehat{\Delta d}}_k[\cos(\tfrac{1}{2}{\widehat{\Delta \theta}_k})\cos\theta_{k}\cos\tilde{\theta}_{k}
    - \cos(\tfrac{1}{2}{\widehat{\Delta \theta}_k})\sin\theta_{k}\sin\tilde{\theta}_{k} \nonumber\\
  & - \sin(\tfrac{1}{2}{\widehat{\Delta \theta}_k})\cos\theta_{k}\sin\tilde{\theta}_{k}
    - \sin(\tfrac{1}{2}{\widehat{\Delta \theta}_k})\cos\tilde{\theta}_{k}\sin\theta_{k}],
\end{align}
where, like our earlier considerations, we assume that \(\cos \left(\tfrac{1}{2}{\widehat{\Delta \theta}_k}\right) \approx 1\), and \(\sin \left(\tfrac{1}{2}\widehat{\Delta \theta}_k\right) \approx \tfrac{1}{2}\widehat{\Delta \theta}_k\). Taking expectations of both sides, we obtain:
\begin{align}
\mathbf{E}[\widehat{\Delta x}_k] & \approx
c_4\mathbf{E}[{\widehat{\Delta d}}_k] \cos \theta_{k} -\tfrac{c_4}{2L}\mathbf{E}[{\widehat{\Delta d}}_k^2]\mathbf{E}[\tan \widehat{\phi}_k]\sin \theta_{k},
\end{align}
where $c_4$ denotes $\mathbf{E}\big[\cos \big(\widetilde{\theta}_{k}\big)\big]$.
Similarly, we obtain
\begin{align}
\mathbf{E}[\widehat{\Delta y}_k]\approx
c_4\mathbf{E}[{\widehat{\Delta d}}_k] \sin (\theta_{k})+\frac{c_4}{2L}\mathbf{E}[{\widehat{\Delta d}}_k^2]\mathbf{E}[\tan (\widehat{\phi}_k)]\cos (\theta_{k}).
\end{align}

\section{Simulations}
\label{sec:simulation}
In this section, we run a number of simulations to evaluate the viability of the developed covariance analysis. The simulations involved calculating the overall cumulative covariance of the vehicle's position and orientation. Comparisons are made with the covariance expressions as reported in Mirats Tur \textit{et al.} \cite{Borja2005}. The following expressions, as reported in Mirats Tur \textit{et al.} \cite[page 1019]{Borja2005}, were used for comparison in our simulation studies:
\begin{subequations}
\begin{alignat}{6}
& \mathbf{E}[\widehat{\Delta x}_k^2]=k_1 \cos ^2(\theta_{k})-k_2 \sin (\theta_{k}) \cos (\theta_{k}) \tan (\phi_k+\widetilde{\phi}^{\max }) \\
& \mathbf{E}[\widehat{\Delta x}_k\widehat{\Delta y}_k]=\tfrac{k_1}{2} \sin (2 \theta_{k})+\tfrac{k_2}{2} \cos (2 \theta_{k}) \tan (\phi_k+\widetilde{\phi}^{\max }) \\
& \nonumber\mathbf{E}[\widehat{\Delta x}_k\widehat{\Delta \theta}_k]=\tfrac{k_1}{L} \cos (\theta_{k}) \tan (\phi_k+\widetilde{\phi}^{\max })\\ & \quad\quad\quad\quad-\tfrac{k_2}{2 L} \sin (\theta_{k}) \tan ^2(\phi_k+\widetilde{\phi}^{\max }) \\
& \mathbf{E}[\widehat{\Delta y}_k^2]=\tfrac{k_1}{2}(1-\cos (2 \theta_{k}))+\tfrac{k_2}{2} \sin (2 \theta_{k}) \tan (\phi_k+\widetilde{\phi}^{\max }) \\
&\nonumber \mathbf{E}[\widehat{\Delta y}_k\widehat{\Delta \theta}_k]=\tfrac{k_1}{L} \sin (\theta_{k}) \tan (\phi_k+\widetilde{\phi}^{\max })\\ &\quad\quad\quad\quad +\tfrac{k_2}{2 L} \cos (\theta_{k}) \tan ^2(\phi_k+\widetilde{\phi}^{\max }) \\
& \mathbf{E}[\widehat{\Delta \theta}_k^2]=\tfrac{k_1}{L^2} \tan ^2(\phi_k+\widetilde{\phi}^{\max })
\end{alignat}
\end{subequations}
where 
\(k_1=\Delta d_k^2+\sigma_{\Delta d}^2\), and 
\(k_2=\frac{1}{L}(\Delta d_k^3+3 \,\Delta d_k\,\sigma_{\Delta d}^2)\). The original work did not provide any explanation of the involved quantity $\widetilde{\phi}^{\max }$. The following (likely to be incorrect) expressions, however, were claimed in \cite[page 1021]{Borja2005}:\footnote{In our simulation studies, we consider \(\widetilde{\phi}^{\max }=0\).}
\begin{subequations}
\begin{alignat}{2}
    \mathbf{E}[\tan (\widehat{\phi}_k)]=& \tan(\phi_k+\widetilde{\phi}^{\max })\\    \mathbf{E}[\tan^2 (\widehat{\phi}_k)]=& \tan ^2(\phi_k+\widetilde{\phi}^{\max })
\end{alignat}\label{EqTanIncorrect}
\end{subequations}
In the Erratum \cite{Borja2007}, it was admitted that the expressions (\ref{EqTanIncorrect}) were not accurate, and based on the physical consideration of depicting the worst-case scenario. 

We conduct a simulation to compute the pose uncertainty, we consider a bicycle motion in a cycloid described by the equations:
\(x(t) = R\sin(2\pi f_0t)\) and 
\(y(t) = R\sin(2\pi f_1t)\),
where we assume $R=10$ m, $f_0=1/100$ and $f_1=1/50$. The random changes in speed and steering angles are assumed to be zero-mean normal. In a total of $200,000$ independent runs, MC simulation was performed by averaging the covariances for each sample time. This MC simulation is then compared with the proposed covariance matrix formulation. 
For a discrete-time simulation, we assume that the time-step is $T=0.1$ sec. Replacing $t$ with $kT$, we obtain: 
\begin{align*}
  x_k = R\sin(2\pi f_0 kT),~~\text{and}~~ y_k = R\sin(2\pi f_1 kT).
\end{align*}
Moreover, the discrete-time orientation is obtained as \[\theta_k=\operatorname{\text{atan2}}(v_{y,k},v_{x,k}),\] where $v_{x,k}$ and $v_{y,k}$ are Cartesian velocities along $x$ and $y$ axes, respectively; they are obtained as 
\begin{align*}
  & v_{x,k} = 2\pi Rf_0\cos(2\pi f_0 kT),\\
  \text{and}~~& v_{y,k} = 2\pi Rf_1\cos(2\pi f_1 kT).
\end{align*}
Given the present pose  $(x_k,y_k,\theta_k)$ and the future pose  $(x_{k+1},y_{k+1},\theta_{k+1})$, we may obtain the desired linear and angular velocities, $v_k$ and $\omega_k$, for time \(t\in[kT,(k+1)T)\) using the inverse kinematics assuming \textit{velocity motion model} as shown below:
\begin{align*}
\mu & =\tfrac{1}{2}\left(\dfrac{\sin (\theta_k)\left(y_{k+1}-y_k\right)+\cos (\theta_k)\left(x_{k+1}-x_k\right)}{\cos(\theta_k)\left(y_{k+1}-y_k\right)-\sin (\theta_k)\left(x_{k+1}-x_k\right)}\right), \\
x_\ast & =\tfrac{1}{2}\left(x_{k+1}+x_k\right)+\mu\left(y_k-y_{k+1}\right), \\
y_\ast & =\tfrac{1}{2}\left(y_{k+1}+y_k\right)+\mu\left(x_{k+1}-x_k\right), \\
\omega_k & =\textstyle\frac{1}{T}\cdot\operatorname{\text{wraptopi}}\big(\operatorname{\text{atan2}}\left(y_\ast-y_k, x_k-x_\ast\right)\\ &-\operatorname{\text{atan2}}\left(y_\ast-y_{k+1}, x_{k+1}-x_\ast\right)\big), \\
\text{and}~~v_k & = \omega_k \sqrt{\left(x_k-x_\ast\right)^2+\left(y_k-y_\ast\right)^2}.
\end{align*}
\footnote{Note that $\operatorname{\text{atan2}}(\mathrm{Y},\mathrm{X})$ is the four quadrant arc-tangent of the elements of $\mathrm{X}$ and $\mathrm{Y}$ such that \(-\pi \le \operatorname{\text{atan2}}(\mathrm{Y},\mathrm{X}) \le \pi\). Also note that $\operatorname{\text{wraptopi}}(\lambda)$ wraps angles in $\lambda$, in radians,
    to the interval $[-\pi, \pi]$ such that $\pi$ maps to $\pi$ and $-\pi$ maps to
    $-\pi$.  In general, odd, positive multiples of $\pi$ map to $\pi$ and odd,
    negative multiples of $\pi$ map to $-\pi$.}where $(x_\ast,y_\ast)$ is coordinates of instantaneous center of rotation, and $\mu$ is  an auxiliary scalar variable; interested readers may refer to \cite{ThrunBook2005} for a detailed description of velocity motion model. So, once the linear and angular velocities, $v_k$ and $\omega_k$, are known, we may compute $\Delta d_k$ and $\phi_k$ as follows:
    \begin{align*}
      \Delta d_k=Tv_k,~~\text{and}~~\phi_k=\operatorname{\text{atan2}}\left(L\omega_k,v_k \right).
    \end{align*}

\begin{figure}[htbp]
   \hspace{-2mm}
\includegraphics[scale=0.09]{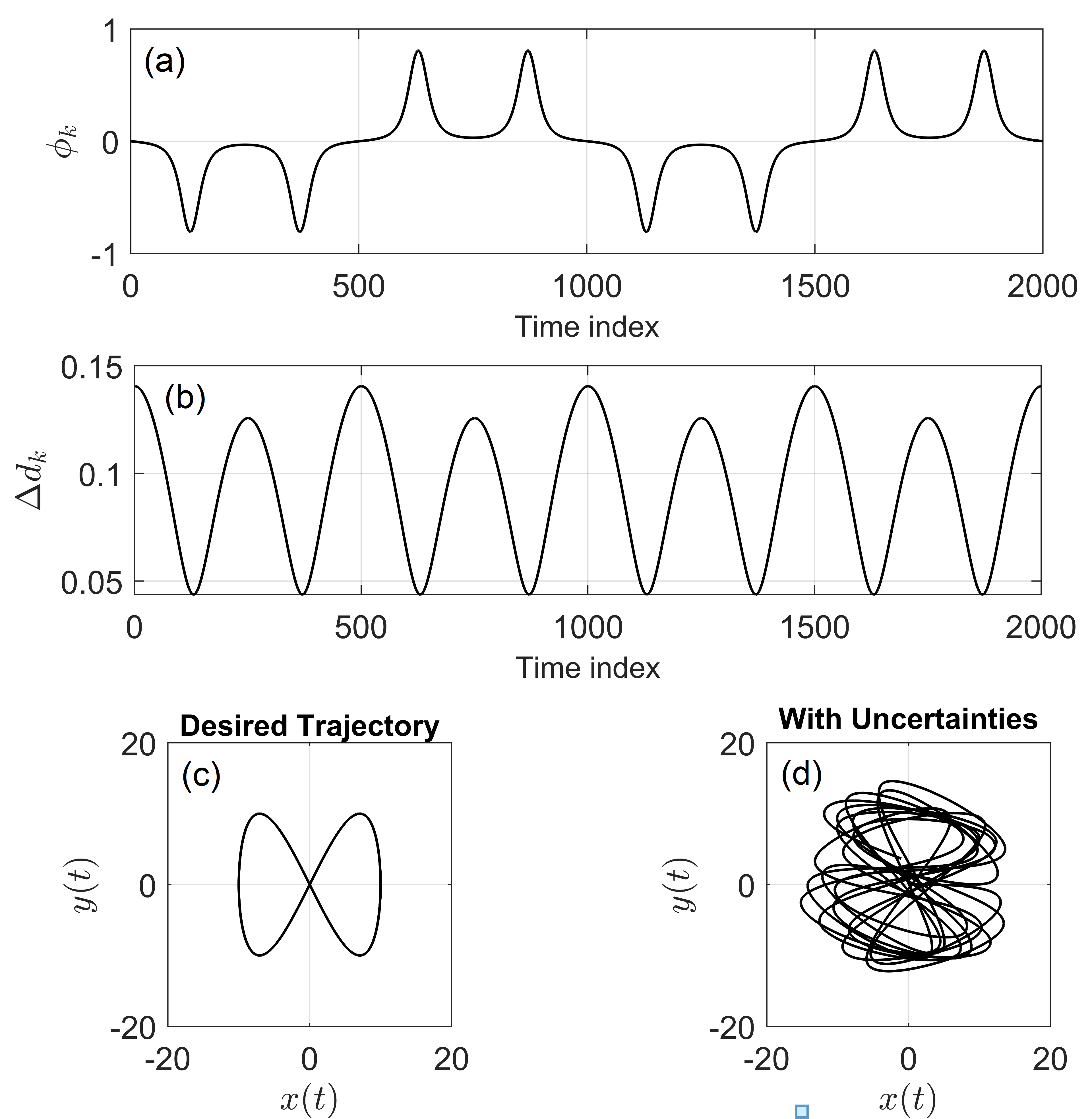}
\caption{(a) Trace of $\phi_k$ in radian for $t\in[0,200]$ sec, (b) trace of $\Delta d_k$ in the meter for $t\in[0,200]$ sec, (c) a round-trip of motion without uncertainties, and (d) motion with uncertainties for $t\in[0,1000]$ sec.} 
\label{fig:fig3}
\end{figure}

\begin{figure*}[htbp]
   \centering
\begin{tabular}{cc}
\includegraphics[scale=0.6]{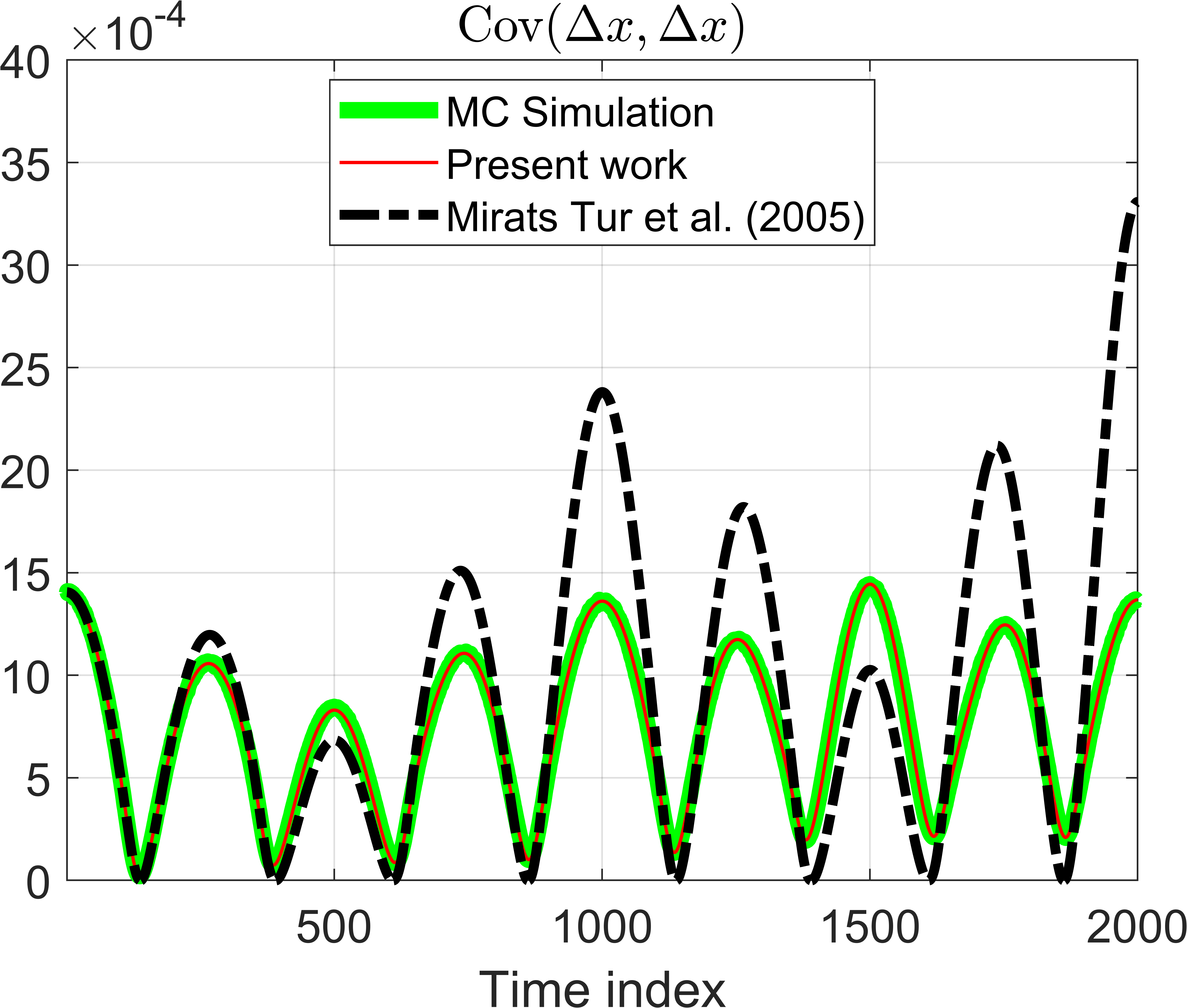} & \includegraphics[scale=0.6]{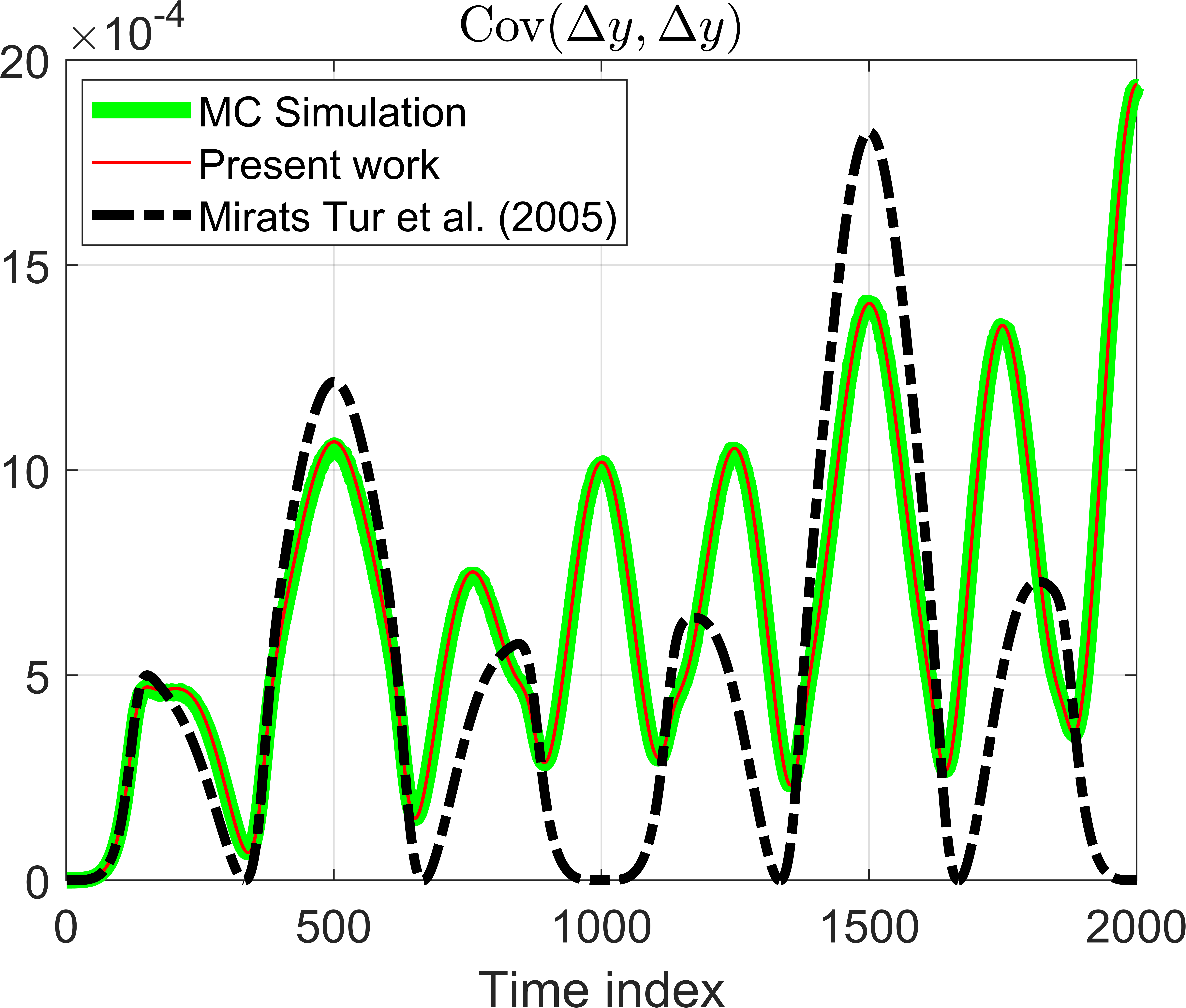}\\ \\ 
\includegraphics[scale=0.6]{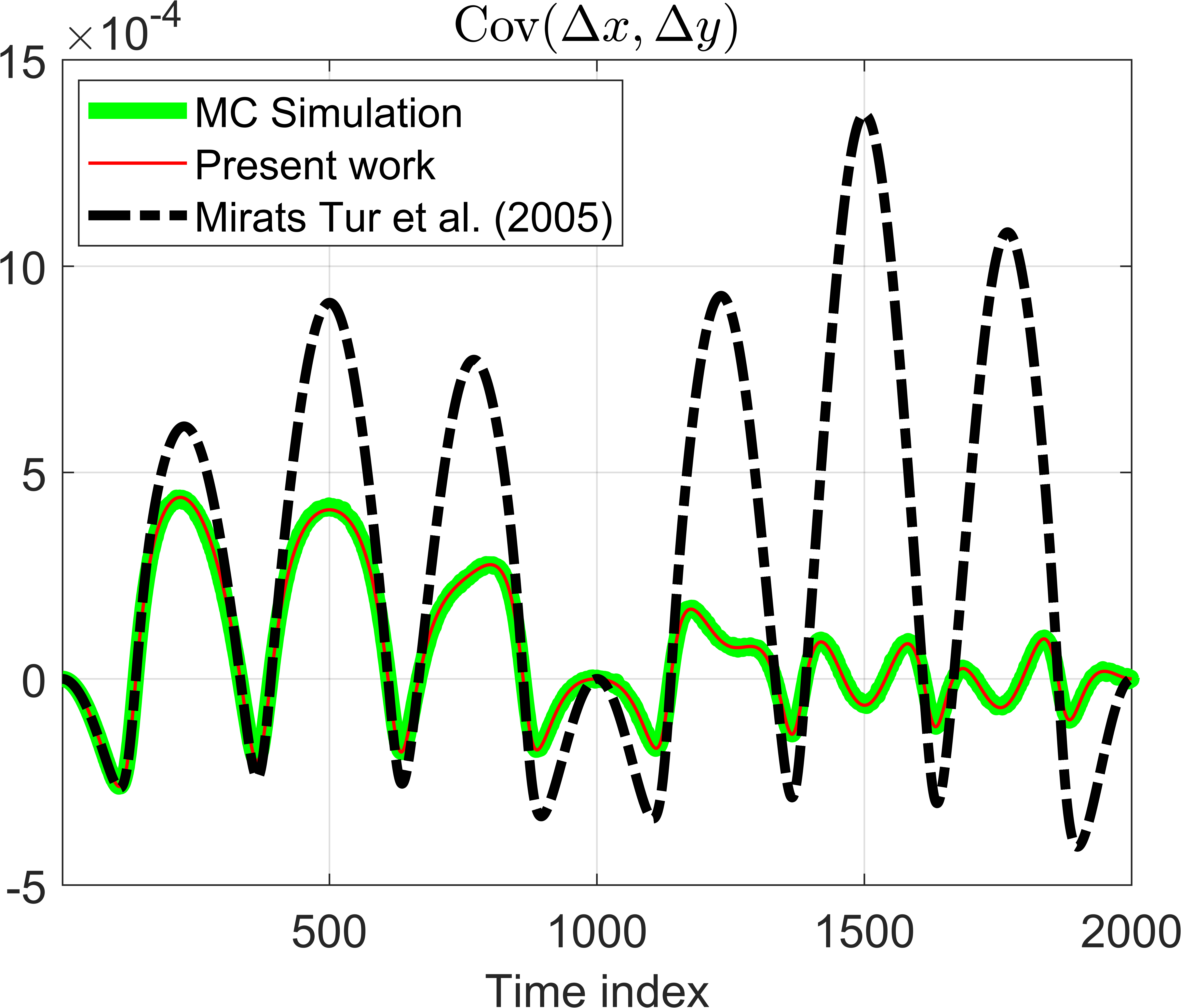} &  \includegraphics[scale=0.6]{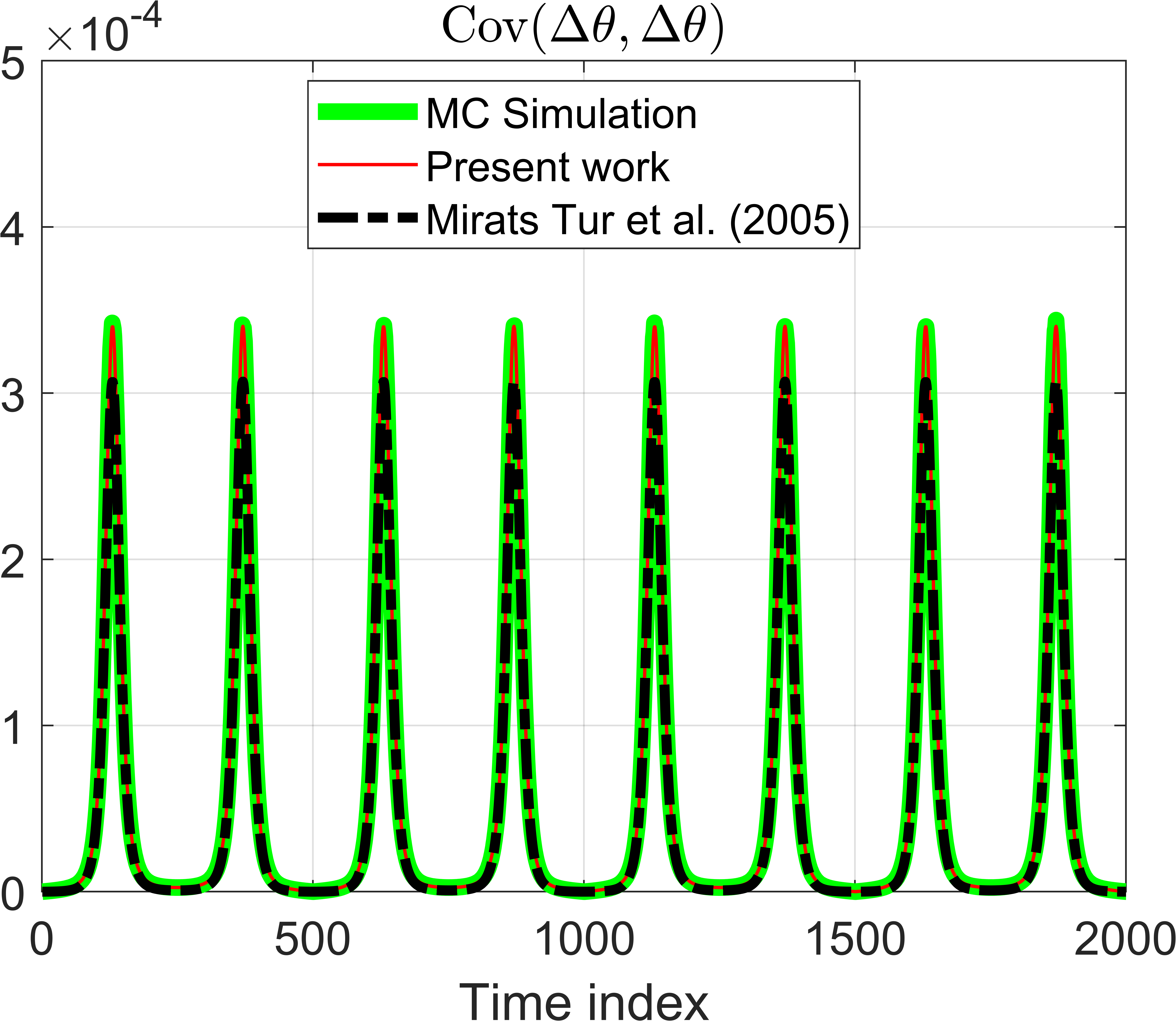}\\ \\ 
\includegraphics[scale=0.6]{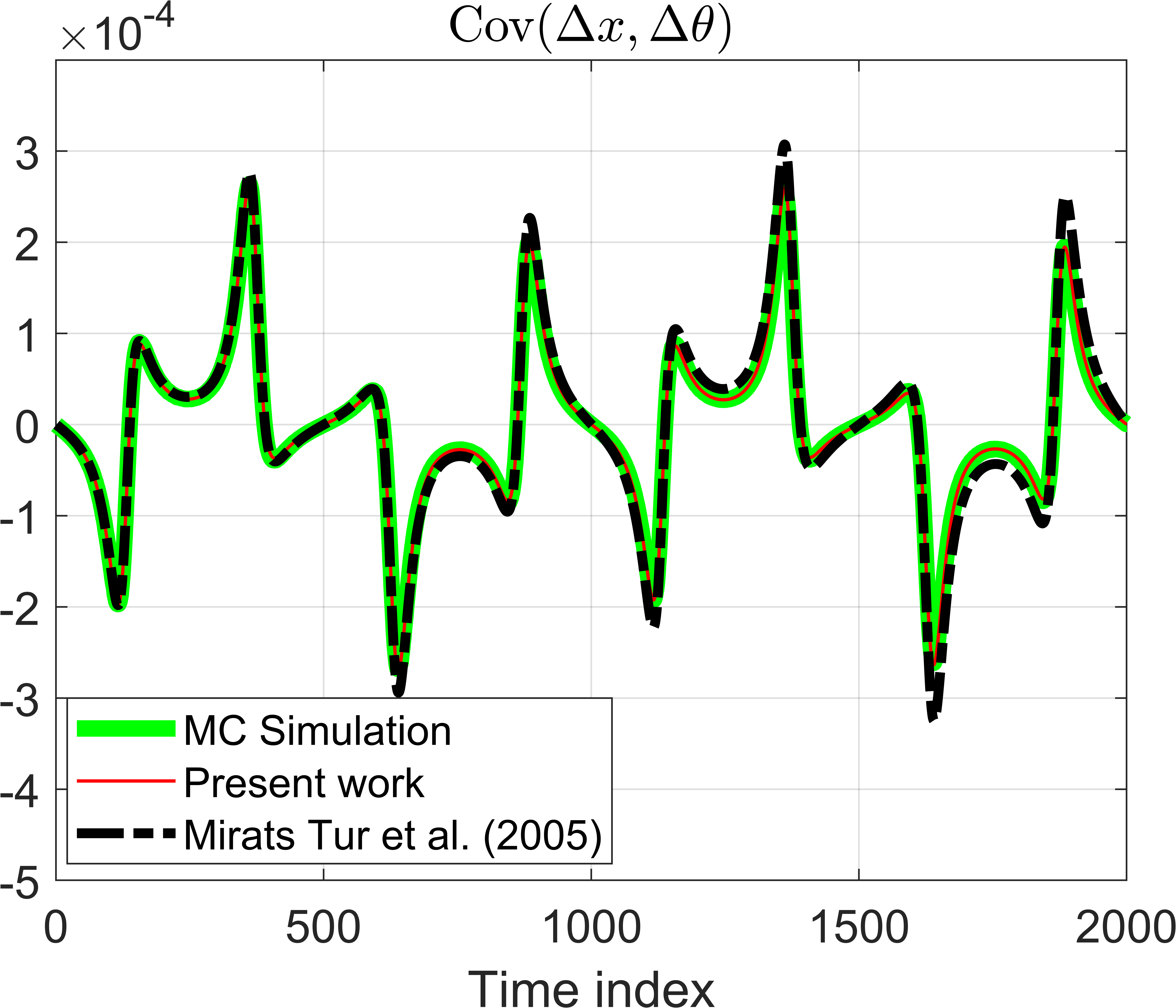} &  \includegraphics[scale=0.6]{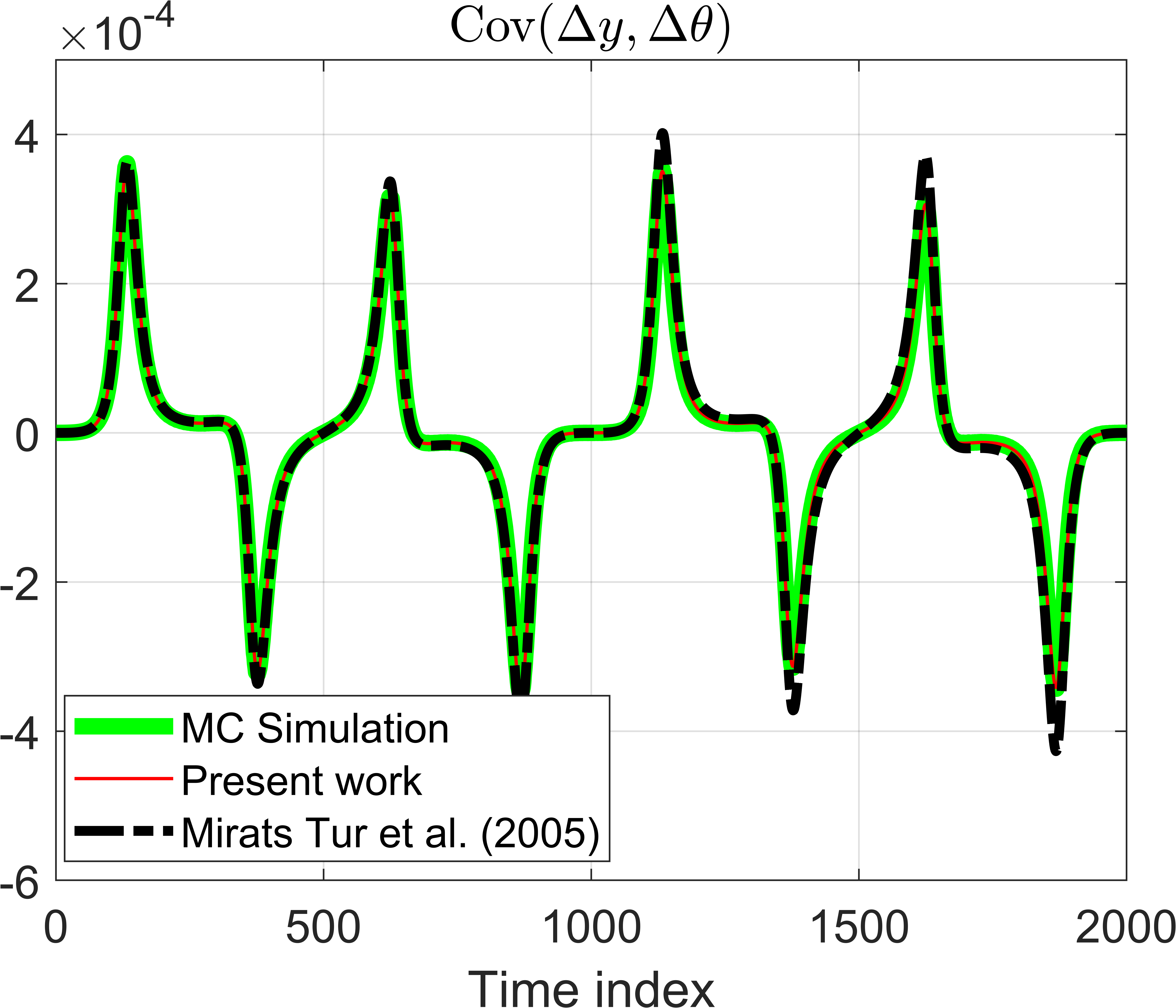}
\end{tabular}
    \caption{Comparison for the covariance values of the different variables of the pose increment $(\Delta x, \Delta y, \Delta \theta)$ while traversing on a \textbf{cycloid path}. The bold green line is a Monte Carlo simulation, and the solid black line is the presented formulation. The values are  $\sigma_{\Delta d}^2=k_d|\Delta d_k| \mathrm{~m},  \sigma_{\phi}^2=k_\phi|\phi_k|, L=1.25 \mathrm{~m}, k_d=0.01, k_\phi=0.005$, number of steps $=2,000$.}
    \label{fig:fig4}
\end{figure*}

Without uncertainty in $\phi_k$ and $\Delta d_k$, the required values of  $\phi_k$ and $\Delta d_k$  as obtained by inverse kinematics (to keep the bicycle on cycloid) are illustrated in 
Fig.~\ref{fig:fig3}(a)-(b) for time $t\in[0,200]$ sec; the cycloid is depicted in Fig.~\ref{fig:fig3}(c). 
Next, we consider a realistic model of uncertainty in $\Delta d_k$ and $\phi_k$, owing to which the random variations in $\Delta d_k$ and $\phi_k$ are proportional to their magnitudes (as suggested in \cite[Chapter 5]{SiegwartBook2011} and \cite{JingdongCUP2015}); therefore, the variances of $\epsilon_{\Delta {d\,k}}$ and $\epsilon_{{\phi\,k}}$, at time index $k$, are obtained as follows: 
\begin{align*}
 \sigma_{\Delta d\,k}^2&=k_d|\Delta d_k|,~~\epsilon_{\Delta {d}\,k} \sim \mathcal{N}\big(0, \sigma_{\Delta d\,k}^2\big)\\   
\sigma_{\phi\,k}^2&=k_\phi|\phi_k|,~~\epsilon_{\phi\,k} \sim \mathcal{N}\big(0, \sigma_{\phi\,k}^2\big)
\end{align*}
where, in this study, we consider $k_d=0.01$ and $k_\phi=0.005$. A single realization of the trajectory of cycloid with uncertainties in consideration is shown in Fig.~\ref{fig:fig3}(d) for time $t\in[0,1000]$ sec. Results of covariance matrix evolution are summarized in Fig.~\ref{fig:fig4}; it may be noticed that the covariance matrix entities obtained in this work are in good agreement with those obtained from MC simulations while the expressions reported in \cite{Borja2005} are unable to capture the correct evolution of covariance matrix components; though, some quantities are close with those of MC simulations while the others may be seen to be largely over-estimated.

\section{Conclusions}

In this work, we have presented statistical analysis to evaluate the covariance matrix for the pose estimation of a nonholonomic bicycle drive system. Relying on a particular discrete-time kinematic model approximation for the vehicle under consideration and an error model, the analysis aimed to obtain a precise closed-form representation of the vehicle's spatial (pose) uncertainties due to imperfect odometric information of wheel displacement and steering angle. The theoretical findings are in good agreement with Monte-Carlo-based computer simulations where the vehicle is moved on a variety of trajectories assuming a Gaussian distributed measurement model. These findings have practical applications beyond academia. They can be applied to state estimation problems in the following ways: (a) state and measurement predictions in pose estimation; (b) motion and path planning; and (c) statistically quantifying errors in odometry.

\bibliographystyle{unsrt}
\bibliography{bibliography_covariance}

@article{Borenstein1996,
  title={Measurement and correction of systematic odometry errors in mobile robots},
  author={Borenstein, J. and Feng, L.},
  journal={IEEE Trans. Robotics and Automation},
  volume={12},
  number={6},
  pages={869--880},
  year={1996},
  publisher={IEEE}
}

@inproceedings{brooks1985visual,
  title={Visual map making for a mobile robot},
  author={Brooks, R.},
  booktitle={Proceedings. 1985 IEEE International Conference on Robotics and Automation},
  volume={2},
  pages={824--829},
  year={1985},
  organization={IEEE}
}

@inproceedings{chatila1985position,
  title={Position referencing and consistent world modeling for mobile robots},
  author={Chatila, R. and Laumond, J.},
  booktitle={Proceedings. 1985 IEEE International Conference on Robotics and Automation},
  volume={2},
  pages={138--145},
  year={1985},
  organization={IEEE}
}

@article{RandallCheeseman1986,
  title={On the representation and estimation of spatial uncertainty},
  author={Smith, R.C. and Cheeseman, P.},
  journal={The International Journal of Robotics Research},
  volume={5},
  number={4},
  pages={56--68},
  year={1986},
  publisher={Sage Publications Sage CA: Thousand Oaks, CA}
}

@incollection{SmithandCheeseman,
  title={Estimating uncertain spatial relationships in robotics},
  author={Smith, R.C. and Self, M. and Cheeseman, P.},
  booktitle={Autonomous Robot Vehicles},
  pages={167--193},
  year={1990},
  publisher={Springer}
}

@techreport{MingWangJ1986,
    author = {Wang, C.M.},
    title = {Error analysis of spatial representation and estimation of mobile robots: GM Research Publication GMR-5579},
    institution = {General Motors Research Laboratories, Warren, Michigan},
    year = {1986}
}

@techreport{MingWangJ1987,
    author = {Wang, C.M.},
    title = {Location estimation and error analysis for an autonomous mobile robot: GM Research Publication GMR-5897},
    institution = {General Motors Research Laboratories, Warren, Michigan},
    year = {1987}
}

@article{MingWangJ1989,
  title={Error analysis of spatial representation and estimation of mobile robots},
  author={Wang, C.M.},
  journal={Communications in Statistics - Theory and Methods},
  volume={18},
  number={3},
  pages={1107--1122},
  year={1989},
  publisher={Taylor \& Francis}
}

@inproceedings{MingWangC1988,
  title={Location estimation and uncertainty analysis for mobile robots},
  author={Wang, C.M.},
  booktitle={Proceedings. 1988 IEEE International Conference on Robotics and Automation},
  pages={1231--1235},
  year={1988},
  organization={IEEE}
}

@article{MingWangJ1990,
  title={Computing uncertainty measures of location estimates for autonomous vehicles},
  author={Wang, C.M.},
  journal={Journal of Statistical Computation and Simulation},
  volume={36},
  number={2-3},
  pages={69--89},
  year={1990},
  publisher={Taylor \& Francis}
}

@techreport{Kleeman1995,
    author = {Kleeman, L.},
    title = {Odometry error covariance estimation for two-wheel robot vehicles: Technical Report MECSE-95-1},
    institution = {Intelligent Robotics Research Centre, Monash University},
    year = {1995}
}

@techreport{ChongKleeman1996,
    author = {Chong, K.S. and Kleeman L.},
    title = {Accurate odometry and error modeling for a mobile robot: Technical Report MECSE-96-6},
    institution = {Intelligent Robotics Research Centre, Monash University},
    year = {1996}
}

@book{Klancar2017,
  title={Wheeled Mobile Robotics},
  author={Klancar, G. and Zdesar, A. and Blazic, S. and Skrjanc, I.},
  year={2017},
  publisher={Butterworth-Heinemann}
}

@inproceedings{kelly2003general,
  title={General Solution for Linearized Error Propagation in Vehicle Odometry},
  author={Kelly, A.},
  booktitle={Robotics Research: The Tenth International Symposium},
  pages={545--558},
  year={2003},
  organization={Springer}
}

@article{KellyAlonzo2004,
  title={Linearized error propagation in odometry},
  author={Kelly, A.},
  journal={The International Journal of Robotics Research},
  volume={23},
  number={2},
  pages={179--218},
  year={2004},
  publisher={SAGE Publications}
}

@article{martinelli2002odometry,
  title={The odometry error of a mobile robot with a synchronous drive system},
  author={Martinelli, A.},
  journal={IEEE Trans. Robotics and Automation},
  volume={18},
  number={3},
  pages={399--405},
  year={2002},
  publisher={IEEE}
}

@article{Borja2005,
  title={A closed-form expression for the uncertainty in odometry position estimate of an autonomous vehicle},
  author={Tur, J.M.M. and Gordillo, J.L. and Borja, C.A.},
  journal={IEEE Trans. Robotics},
  volume={21},
  number={5},
  pages={1017--1022},
  year={2005},
  publisher={IEEE}
}

@article{Borja2007,
  title={Erratum to: A closed-form expression for the uncertainty in odometry position estimate of an autonomous vehicle.},
  author={Tur, J.M.M. and Gordillo, J.L. and Borja, C.A.},
  journal={IEEE Trans. Robotics},
  volume={23},
  number={},
  pages={1302-1302},
  year={2007},
  publisher={IEEE}
}

@article{Chapman1970,
  title={Moments, variances, and covariances of sines and cosines of arguments which are subject to random error},
  author={Chapman, J.W.},
  journal={Technometrics},
  volume={12},
  number={3},
  pages={693--694},
  year={1970},
  publisher={Taylor \& Francis}
}

@article{RodriguezArevalo2018,
  title={On the importance of uncertainty representation in active SLAM},
  author={Rodriguez-Arevalo, M.L. and Neira, J. and Castellanos, J.A.},
  journal={IEEE Transactions on Robotics},
  volume={34},
  number={3},
  pages={829--834},
  year={2018},
  publisher={IEEE}
}

@article{KnuthPrabir2013,
  title={Error growth in position estimation from noisy relative pose measurements},
  author={Knuth, J. and Barooah, P.},
  journal={Robotics and Autonomous Systems},
  volume={61},
  number={3},
  pages={229--244},
  year={2013},
  publisher={Elsevier}
}

@article{JingdongCUP2015,
  title={An efficient approach to pose tracking based on odometric error modelling for mobile robots},
  author={Yang, J. and Yang, J. and Cai, Z.},
  journal={Robotica},
  volume={33},
  number={6},
  pages={1231--1249},
  year={2015},
  publisher={Cambridge University Press}
}

@article{FilhoAccess2019,
  title={The impact of parametric uncertainties on mobile robots velocities and pose estimation},
  author={Filho, D.C. and Nunes, J.G. and Carvalho, E.A.N. and Molina, L. and Freire, E.O.},
  journal={IEEE Access},
  volume={7},
  pages={69070--69086},
  year={2019},
  publisher={IEEE}
}

@ARTICLE{Borja2009,
  author={Borja, C.A. and Tur, J.M.M. and Gordillo, J.L.},
  journal={IEEE Robotics \& Automation Magazine}, 
  title={State your position}, 
  year={2009},
  volume={16},
  number={2},
  pages={82-90},
  }

@article{ColinDas2010,
  title={Calibration of kinematic odometric parameters for differential drive mobile robots: An overview},
  author={Das, C.},
  journal={Inf. Tech., University of Illinois in Aerospace Engineering},
  year={2010}
}

@article{Langholm2021,
  title={Banana distributions based on stochastic polar coordinates},
  author={Langholm, T. and Totland, H.},
  year={2021},
  publisher={FHS, Sj{\o}krigsskolen}
}

@book{ThrunBook2005,
    author = {Thrun, S. and Burgard, W. and Fox, D.},
    title = {Probabilistic Robotics},
    publisher = {The MIT Press},
    year = {2006}
}

@book{SiegwartBook2011,
    author = {Siegwart, R. and Nourbakhsh, I.R. and Scaramuzza, D.},
    title = {Introduction to Autonomous Mobile Robots},
    publisher = {The MIT Press},
    year = {2011}
}

\end{document}